%% file: main.tex
\documentclass[sigconf]{acmart}
\setcopyright{none}
\acmConference[HAI '26]{The Fourteenth International Conference on Human-Agent Interaction}{November 16--19, 2026}{Osaka, Japan}
\acmISBN{}
\acmDOI{}
\copyrightyear{2026}
\acmYear{2026}

\usepackage{graphicx}
\usepackage{subcaption}
\usepackage{amsmath}
\usepackage{float}
\usepackage{xcolor}
\usepackage{url}
\usepackage{soul}

\sethlcolor{yellow}
\soulregister\cite7
\soulregister\ref7
\soulregister\pageref7

\copyrightyear{2026}
\acmYear{2026}
\setcopyright{cc}
\setcctype{by-nc-nd}
\acmConference[HAI '26]{Proceedings of the 14th International Conference on Human-Agent Interaction}{November 16--19, 2026}{Osaka, Japan}
\acmBooktitle{Proceedings of the 14th International Conference on Human-Agent Interaction (HAI '26), November 16--19, 2026, Osaka, Japan}
\acmDOI{10.1145/3841580.3841620}
\acmISBN{979-8-4007-2575-3/2026/11}

\begin{document}

\title{Socrates went Nuclear: Comparing Interaction Strategies for AI systems in a Learning Context using Brain Sensing}

\author{Alexandre Clin Deffarges}
\email{aclindeffarg@student.ethz.ch}
\affiliation{%
  \institution{ETH Zürich}
  \city{}
  \state{}
  \country{}
}

\author{Nataliya Kosmyna}
\email{nkosmyna@mit.edu}
\affiliation{%
  \institution{MIT Media Lab}
  \city{}
  \state{}
  \country{}
}

\author{Pattie Maes}
\email{pattie@mit.edu}
\affiliation{%
  \institution{MIT Media Lab}
  \city{}
  \state{}
  \country{}
}

\renewcommand{\shortauthors}{Clin Deffarges, Kosmyna, and Maes}

\begin{abstract}
\input{sections/abstract}
\end{abstract}




\maketitle

\section{Introduction}
\input{sections/introduction}

\section{Related Work}
\input{sections/related_work}

\section{Study Design}

\input{sections/methodology}

\section{Data Analysis Plan}
\input{sections/data_analysis_plan}

\section{Results}
\input{sections/results}

\section{Discussion}
\input{sections/discussion}

\section{Conclusion}
\input{sections/conclusion}

\input{sections/references}
\appendix
\input{sections/appendix}

\end{document}

%% file: sections/abstract.tex
Students are adopting LLMs at scale. This poses a question:
does unrestricted AI access bypass the cognitive effort required for learning,
or does it streamline knowledge acquisition? This paper reports on a study where we compare three designs for user-AI interaction in a learning context:
(1) an unrestricted conversational bot like ChatGPT,
(2) a pedagogically constrained bot that guides through hints
without giving final answers, which we refer to as the Socratic mode;
and (3) a non-conversational adaptive tutoring system
that adjusts difficulty in real-time based on the user's cognitive engagement derived from the brain signals.
Fifty study participants were tasked with learning about nuclear safety protocols, a domain chosen for its zero-prior knowledge baseline.
The participants progressed through an instructional video,
a pre-test, an AI-driven assessment phase, which varied in the three conditions, and an immediate post-test.
The nature of the questions centered primarily on factual knowledge acquisition, but it still required participants to have a global understanding of the concepts in order to answer the questions correctly. 
A Muse headband was used to derive the cognitive engagement of all users in all conditions.

The unrestricted chatbot produced higher learning gains (delta)
than both constrained modes ($p < .03$, $d > 0.80$),
while the adaptive condition generated significantly higher EEG engagement
($p = .018$).
The cluster analysis of chatbot usage and discussion patterns by users showed that most participants in the unrestricted-mode adopted a direct answer-retrieval strategy, while participants in the Socratic-mode initially attempted to reason through the hints before progressively disengaging.
Consequently, this also suggests that the success of the unrestricted AI is not an evidence of deeper learning, but rather a result of the immediate post-test evaluation after the training phase.

%% file: sections/introduction.tex
The way students study and work on their own has changed drastically since 2022.
Across all educational levels, they increasingly turn to general purpose LLMs like ChatGPT
to work through assignments and homework.
Recent survey data indicate that over 89\% of students have used ChatGPT
for homework assistance, 53\% using it to write essays and 48\%
for unproctored assessments.
A 2026 Pew Research Center study found that 54\% of adolescents
actively use AI for schoolwork \cite{pew2026}.

What this means for learning is not clear.
One perspective, based on cognitive load theory and the desirable difficulties framework \cite{fesler2026},
is that general-purpose LLMs might threaten deep learning
because they are built to be helpful and resolve tasks quickly 
and thus may reduce the productive struggle that builds lasting understanding \cite{fesler2026}.
Students seem to sense this themselves:
many adolescents (59\%) see AI-assisted academic dishonesty
as systemic in their institutions,
and 72\% of college students support the ban on ChatGPT
in universities \cite{pew2026}.

However, a competing perspective suggests that unrestricted AI access
may offer underappreciated benefits by providing immediate, high-quality explanations on demand. This perspective posits that
general purpose LLMs enable efficient knowledge acquisition
that can serve as a foundation for deeper understanding.
Unrestricted access also lets students explore tangential questions
and adapt the LLM's output to their own level.
The empirical evidence for which perspective better predicts
actual learning outcomes is very mixed \cite{fesler2026}.

In response to these concerns, two directions have emerged.
The first focuses on Socratic conversational LLMs which
are constrained to guide students through questions and hints
rather than giving answers.
Theoretically,  such LLMs have the potential to support deeper cognitive processing.
But challenges include potential frustration over a series of questions and longer time requirements. Additionally, without clear guidance, the Socratic chat could remain superficial \cite{fakour2025}. Additionally, relying on LLMs too much might also hinder critical thinking \cite{fakour2025}.

Motivated by recent papers and a growing interest in personalized adaptive learning \cite{hwang2026}, the second approach focuses on adaptive, non-conversational exercise systems \cite{adaptive_difficulty2025}.
Here, AI analyzes the performance of the user in real-time and
restructures questions based on the success of previous responses of the said user.
There is also a possibility to leverage physiological measurements in order to estimate the learner's engagement or cognitive load
and adapt an output of an LLM accordingly.
This adaptive approach moves beyond the open-ended dialogue of a typical chatbot.

Although prior work, including personalized feedback mechanisms explored in systems like NeuroChat \cite{baradari2025}, demonstrates that personalized interaction improves engagement outcomes, the literature lacks a direct empirical comparison between an unrestricted chatbot, a Socratic bot, and an adaptive bot in the context of homework assistance.

We ran a controlled experiment to compare these three approaches: unrestricted, Socratic and adaptive. To handle the confound of heterogeneous prior knowledge, we chose a topic of nuclear safety protocols, currently rare enough that our participants would not demonstrate any pre-existing knowledge. The experiment simulated an independent study session: participants first watched a standardized video, took a baseline pre-test, and were then randomly assigned to one of the three modes. We administered a post-test immediately following the experiment to measure direct knowledge retention. Beyond test scores, we collected neurophysiological recordings using a Muse 2 EEG headband \cite{muse2} to measure cognitive engagement in all three conditions. Only the adaptive condition used the real time EEG data to modify its content; in the other two conditions, the EEG was recorded passively. This setup allowed us to evaluate both immediate knowledge retention and the use of cognitive engagement to directly influence bot output.

%% file: sections/related_work.tex
The literature on AI in education is large and fragmented. We organize it around six themes that directly inform our study design: the cognitive cost of unrestricted AI use, the theoretical basis for effortful learning, Socratic and constrained architectures, adaptive tutoring systems, and neurophysiological measurement of engagement.

\subsection{The Cognitive Cost of Unrestricted AI in Education}

Several recent studies suggest that using general LLMs during learning has a measurable cognitive cost. Kosmyna, Hauptmann et al.~\cite{kosmyna2025} ran an EEG study with 54 participants across four essay-writing sessions and showed that students relying on ChatGPT showed weaker functional brain connectivity during the task than those using search engines or no tools. Stadler, Bannert and Sailer~\cite{stadler2024} reached a similar conclusion: in a randomized trial with 91 students learning about nanoparticles, LLM users reported lower cognitive load and produced weaker reasoning. Gerlich~\cite{gerlich2025} argued that delegating mental work to AI tools could decrease critical thinking over time.

Large scale surveys documented a similar effect. A RAND report~\cite{rand2025} stated that students themselves think AI is hurting their critical thinking. Fan et al.~\cite{fan2025} described a "metacognitive laziness'' effect: when learners overrely on AI answers, they lose track of what they know and do not know, leading to a false sense of knowledge. Rizvi et al.~\cite{assefa2024} described LLMs as a double-edged sword, and Ortiz-Bonnin and Blahopoulou~\cite{azpiazu2025} showed that ChatGPT usage is blurring the line between help and cheating.

\subsection{The Need for Effort}

Bjork and Bjork~\cite{bjork2011} formalized the idea of desirable difficulties: techniques that make learning harder in the short term but improve long term retention. Effort during learning strengthens memory, which is precisely what AI tools can bypass. Wells et al.~\cite{wells2025} brought these ideas to the current use of AI and explained that productive struggle is potentially what is missing when using an LLM. They also connected this to Vygotsky's Zone of Proximal Development (ZPD) \cite{vygotsky1978}: the difficulty needs to be tailored to the learner's level.

\subsection{Socratic Architectures and Constrained LLM Design}

Since LLMs tend to give direct answers, recent work tries to move  them towards Socratic implementations, meaning that they guide students toward the answer through targeted questions.
Dinucu-Jianu, Macina et al.~\cite{pedagogicalrl2025} used reinforcement learning to reward
"quality" teaching (fosters understanding and reasoning) instead of focusing on correct answers alone.

Several systems targeted specific subjects. Xie et al.~\cite{xie2025} built STAP, a Socratic
programming tutor with adaptive step-by-step support.
Liu et al.~\cite{pace2025} created PACE, a math tutor that simulates each student's learning style and aligns responses with the corresponding persona. They also applied the Socratic method to deliver instant feedback and stimulate deeper reasoning. This enabled PACE to adapt to
individual learner needs.

Cohn et al.~\cite{inquizzitor2025} proposed Inquizzitor, an LLM-based agent for science assessment that scores student responses and delivers adaptive feedback in the Zone of Proximal Development (ZPD, \cite{vygotsky1978}),
reaching high agreement with human graders ($\kappa_w \geq 0.70$).

\subsection{Adaptive Learning Systems and Intelligent Tutoring Systems}

Vanacore et al.~\cite{vanacore2026} argued that LLMs must be combined with established Intelligent Tutoring Systems (ITS, adaptive software that delivers personalized instruction and feedback~\cite{vanlehn2011}) methods to become real tutors rather than generic chatbots. Scarlatos et al.~\cite{scarlatos2025} went further by training LLM tutors to track student knowledge directly within the conversation. Work on adaptive difficulty models~\cite{adaptive_difficulty2025} proposed algorithms that adjust problem complexity in order to keep students in the optimal difficulty range.

\subsection{Neurophysiological Measurements and Neuro-Adaptive Interfaces}

Several recent works connect the learner's cognitive state directly to AI using physiological measurements such as brain signals. Baradari, Kosmyna et al.~\cite{baradari2025} developed NeuroChat, an AI tutor that uses real-time EEG to derive cognitive engagement of the subjects in order to adjust the LLM difficulty. Wearable EEG devices~\cite{wearable_eeg2025} now make this kind of cognitive tracking possible outside the laboratory~\cite{kosmyna2026neuroskill}. Noel et al.~\cite{noel2025} used the same engagement index ($\text{beta} / (\text{theta} + \text{alpha})$) to show how different teaching methods impacted student's focus. Beauchemin et al.~\cite{beauchemin2024} built a system that adjusts learning speed based on cognitive load. Gkintoni et al.~\cite{gkintoni2025} reviewed 103 empirical studies and argued that AI systems combined with neurophysiological monitoring (such as EEG) can dynamically adjust difficulty in real time to keep learners engaged.

Putting this together, unrestricted LLMs can undercut the cognitive effort that contributes to learning. Socratic prompting and adaptive tutoring each address this problem to an extent, but have only been studied separately, and while physiological sensing like EEG can track engagement in real time, it has rarely been used to compare these fundamentally different AI tutoring styles. To the best of our knowledge, no studies have been reported that compare an unrestricted chatbot, a Socratic bot, and an EEG-adaptive exercise system side by side under controlled conditions while measuring both learning and engagement in all three conditions.

%% file: sections/methodology.tex
To understand how these different AI tools actually impact learning and student engagement, we ran a controlled study. This section explains the experimental design, the three AI modes compared, and how brain activity was measured.

\subsection{The Three AI Interaction modes}

\begin{figure}[htbp]
\centering
\begin{subfigure}[t]{0.48\columnwidth}
    \centering
    \includegraphics[width=\linewidth]{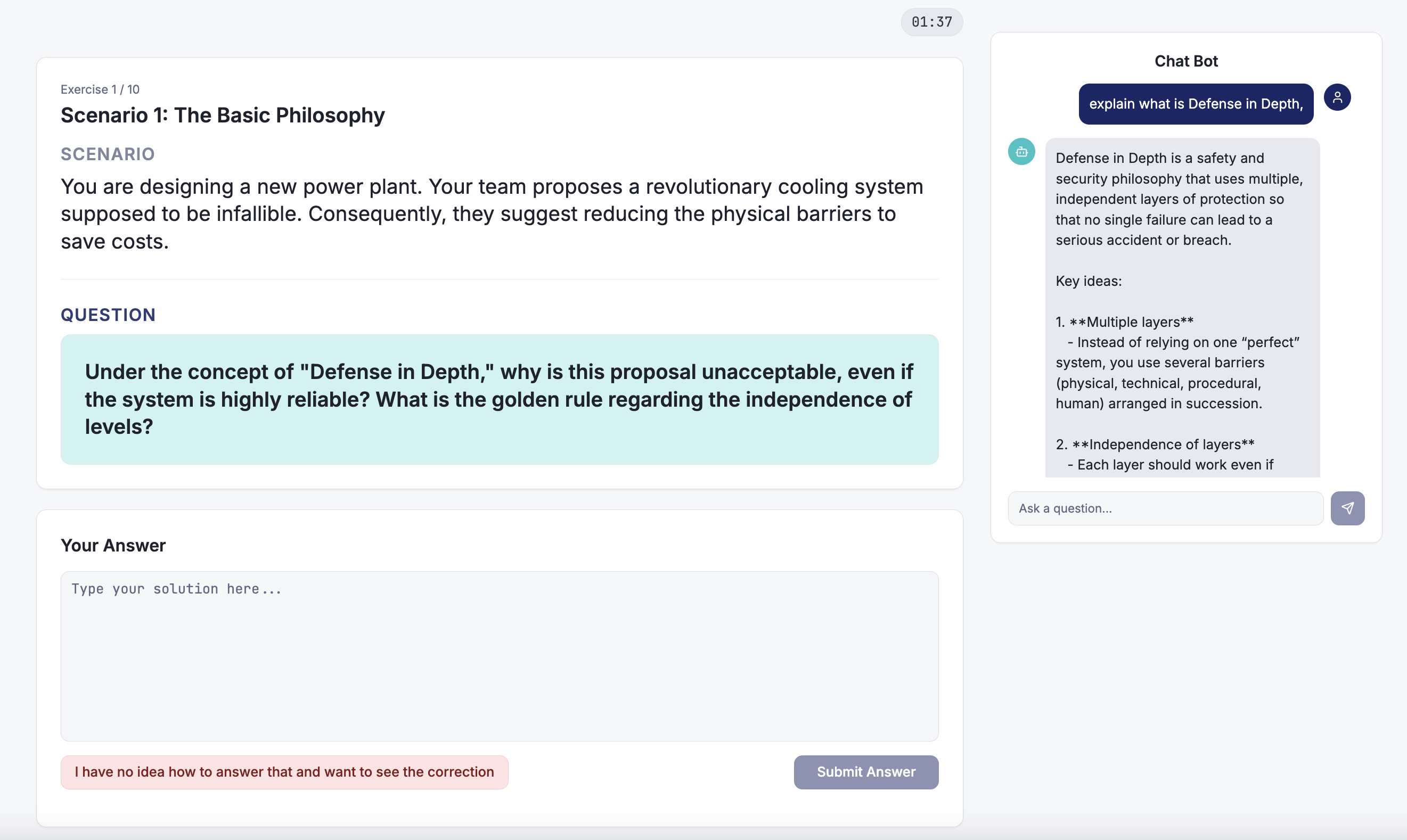}
    \caption{Modes 1 and 2: chatbot on the side.}
    \label{fig:screenshot_chatbot}
\end{subfigure}
\hfill
\begin{subfigure}[t]{0.48\columnwidth}
    \centering
    \includegraphics[width=\linewidth]{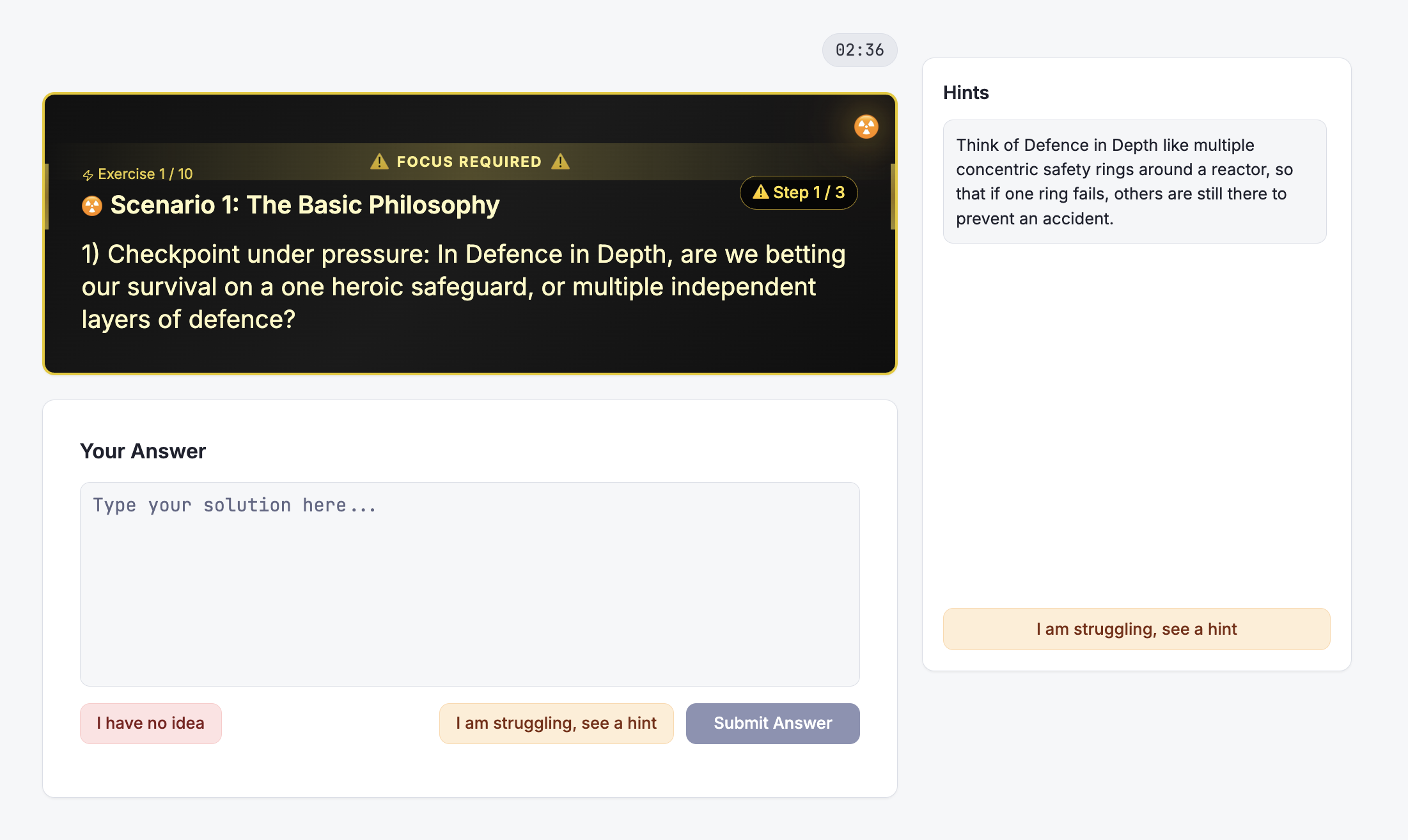}
    \caption{Mode 3: EEG-driven feedback.}
    \label{fig:screenshot_mode3_gamified}
\end{subfigure}
\caption{Interface screenshots. (a)~Modes 1--2 share the chatbot-on-the-side layout. The screenshot shows Mode~1; Modes 1 and 2 share the exact same interface on purpose (only the system prompt changes), so a Mode~2 screenshot would look identical. (b)~In Mode~3, EEG-driven feedback adds a ``Focus Required'' alert and high-contrast elements when low engagement is detected.}
\Description{Two screenshots side by side showing the conversational interface and the Mode 3 interface.}
\end{figure}

\begin{itemize}
    \item \textbf{Mode 1 : Baseline, Unrestricted Chatbot :} A standard ChatGPT-like bot integrated in a side panel to the right of the problem interface, without any pedagogical restrictions (see Figure~\ref{fig:screenshot_chatbot}). If the student asks for the final answer, the AI gives it directly. EEG engagement is derived and recorded passively and continuously.
    
    \item \textbf{Mode 2 : Socratic, Restricted Chatbot :} The exact same interface as in Mode 1 (see Figure~\ref{fig:screenshot_chatbot}), but the side bot is constrained by a Socratic tutoring system prompt inspired by the tutoring prompts of Tack and Piech \cite{tack2023}, but rewritten for this study (the full prompt is provided in Appendix~\ref{app:prompt}). It never gives the final answer. Instead, it asks guiding questions and provides hints to nudge the student in the right direction. The pedagogical policy goes beyond simply hiding the answer: at every turn, the bot compares the student's current answer with the reference correction to identify what is missing or wrong, asks one guiding question at a time, consolidates progress by asking the student to restate ideas in their own words, and ends every message with a short directional hint. The theoretical rationale is that hiding the answer while providing the guidance step-by-step keeps the student working just above their current level, in the zone of proximal development (ZPD) \cite{vygotsky1978}, while the one-question / one-hint format keeps mental load low so that the effort goes into understanding \cite{sweller1998,sweller2010}. EEG engagement is also derived and recorded passively and continuously.
    \item \textbf{Mode 3 Adaptive Exercise AI: No Chatbot:} Here we do not have a conversational interface. If the student fails (< 80\%, denoting a partial or incorrect answer based on the evaluation criteria), the system redirects them through a set of simpler sub-questions. If they also struggle with those, they can generate tailored hints for each one. In addition, this mode actively ingests real-time EEG data to adapt visual feedback and question framing. If the engagement drops below a baseline (calculated for each person), the interface introduces visual stimuli and reformulates the question. Low engagement paired with incorrect answers triggers strong visual feedback (e.g., explosion animations, punchy dramatic questions, etc; see Figure~\ref{fig:screenshot_mode3_gamified} and Figure 2). Correct answers with high engagement get positive reinforcement.
\end{itemize}

The instructional inputs were identical across the three modes: the same video, ten conceptual topics, and AI always had access to the problem statement and the reference correction. What differed was only how much of that information the system was allowed to reveal. Mode 1 participants, therefore, effectively had easier access to more information, and this was deliberate: an unrestricted chatbot that gives direct answers is exactly what students face when they do their homework with a general-purpose LLM. Likewise, the sub-question redirection is specific to Mode 3 by design: Modes 1 and 2 are about participants thinking through the problem themselves, the AI is there, but the thinking still has to come from them, not from the bot, whereas Mode 3 is different: the LLM guides them directly, using sub-questions to lay out the steps toward the answer. Giving those same sub-questions to Modes 1--2 would have mixed the conditions we were trying to compare.

\begin{figure}[htbp]
\centering
\includegraphics[width=0.6\columnwidth]{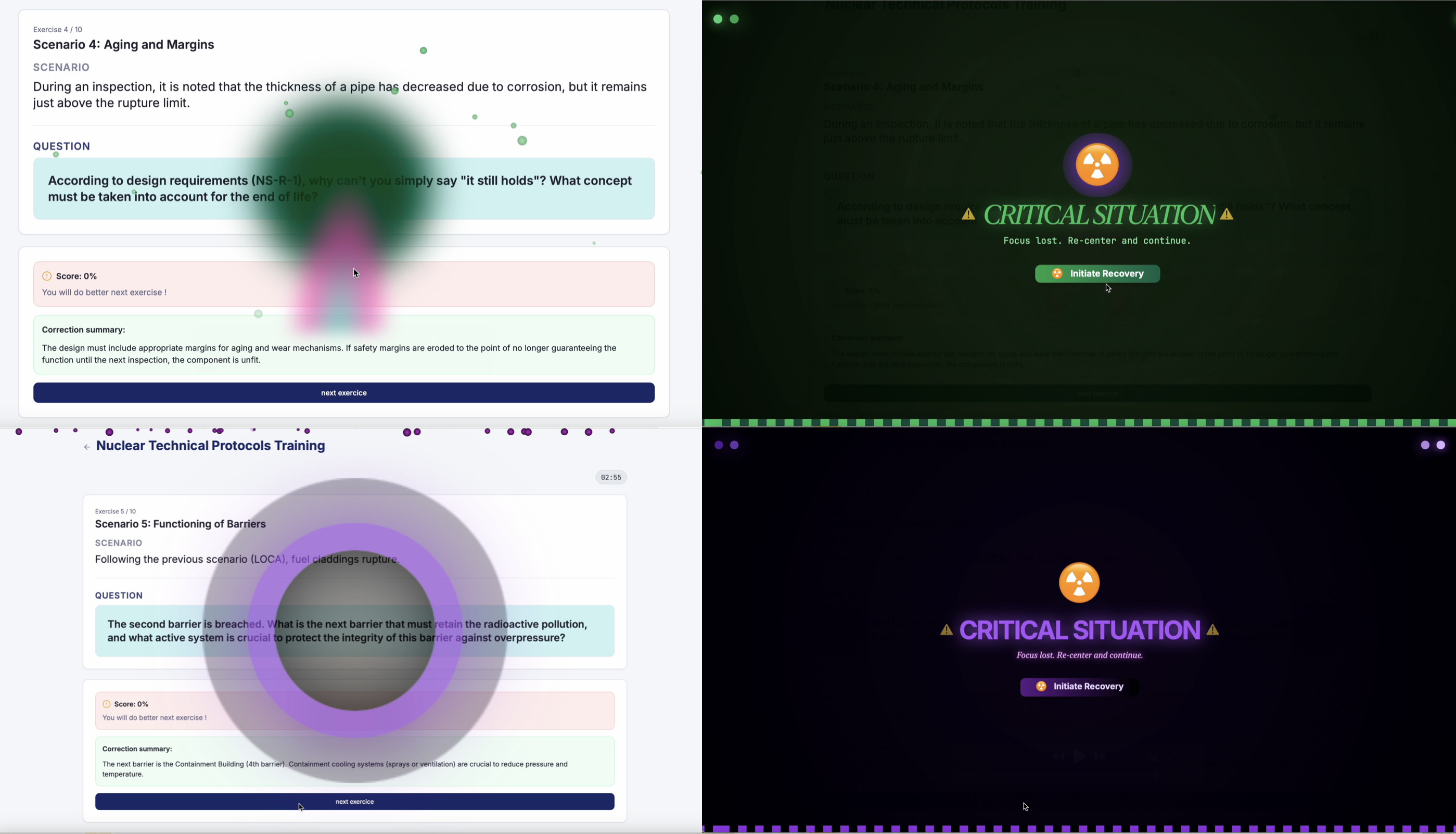}
\caption{Examples of the EEG-triggered visual feedback (explosion) in Mode 3.}
\Description{Examples of the EEG-triggered visual feedback (explosion) in Mode 3.}
\label{fig:screenshot_mode3_explosion}
\end{figure}

\textbf{Common evaluation system.} For each mode, answers are evaluated in real time using a prompt that includes the problem statement, the student's answer, and the problem correction. Students must score at least 80\% to pass. Otherwise, they can try again as many times as needed and receive brief explanatory feedback. Pilot testing demonstrated that an 80\% score reliably indicates conceptual understanding, whereas 100\% typically requires exact specific vocabulary.

\textbf{Conversational AI backend (Modes 1 and 2).} Both conversational Modes (mode 1 and 2) use the OpenAI GPT-5.2 model. The chatbot has access to the current problem statement, the student's answer so far and past trials, and the reference correction. For Mode 2, the system prompt is inspired by the tutoring prompt from Tack and Piech \cite{tack2023}, but was rewritten for this study (full text in Appendix~\ref{app:prompt}); it enforces Socratic pedagogical constraints: the model never provides the final answer and instead guides the student through subquestions, hints, and targeted challenges.

\subsection{Hypothesis}

Based on prior research, we formulate three hypotheses. \textbf{(H1) The Socratic method leads to better learning:} participants using the Socratic Chatbot (mode 2) will achieve higher learning deltas ($\Delta = \text{post-test} - \text{pre-test}$) compared to the Unrestricted Chatbot (mode 1), because the deeper cognitive processing required is expected to strengthen retention. The adaptive mode (mode 3) may also produce superior learning delta, though this prediction is more uncertain. \textbf{(H2) Real-time adaptation maximizes cognitive engagement:} participants in the adaptive mode will show higher cognitive engagement (measured by Beta / (Alpha + Theta) \cite{pope1995, baradari2025}) than the conversational conditions, thanks to dynamic difficulty adjustment in the zone of proximal development \cite{vygotsky1978} and real-time EEG-driven visual feedback. \textbf{(H3) Socratic and adaptive modes enhance perceived engagement:} participants in the adaptive (mode 3) and Socratic (mode 2) conditions will report greater subjective engagement and perceived learning, and may also report higher perceived difficulty.

\subsection{Experimental Procedure}

To ensure a fair and consistent comparison across all 3 Modes, every participant went through the exact same seven-step study session.

\textbf{Experimental Phases:} (1) Background questionnaire (with prior knowledge self-assessment); (2) EEG calibration (resting state + Stroop task); (3) Learning phase (10 minutes video); (4) Baseline assessment (pre-test: 10 open-ended questions); (5) AI-assisted study session (randomized mode, 10 open-ended questions with assistance and grading); (6) Final assessment (post-test, 10 open-ended questions); (7) Post-experiment questionnaire.

The experiment took place in a dedicated study room of the research lab on the institute's campus. Participants faced a wall sitting at an empty table, free from any external distractions like posters, etc. The room is a closed space, without windows. The verbal instruction script at the beginning of the experiment was identical for all participants.

There were no strict time limits, participants received recommended durations (~10 minutes for the pre and post test, ~30 minutes for the study session) with a timer on the  screen to help them pace themselves. Time on task was not equal across conditions, due to the nature of the conditions themselves, since a tutor that never provides a direct answer and proceeds by Socratic questioning, will necessarily takes longer. We had to choose between sacrificing equal time or an equal number of questions, and we preferred to hold the number of questions constant, both to simulate a real homework situation (where students face a fixed set of exercises rather than a fixed duration of time) and because it makes per-question comparisons more direct.

Mode 3 participants did not receive any hands-on practice phase with the EEG-triggered feedback before training: they only received a short standardized verbal explanation of focus alerts and of how the EEG feedback works.

\textbf{EEG Protocol and Calibration.} The calibration consisted of two phases: first, 30 seconds with eyes open in a resting state (staring at the wall), establishing a low-engagement baseline; second, 30 seconds doing the Stroop Color-Word Test, where colored shapes appear with color words printed in misleading ink colors and participants must pick the word that actually matches the shape's color. The Stroop task induces high cognitive load \cite{stroop1935}, providing $E_{\min}$ and $E_{\max}$ values for future normalization. Engagement was also passively measured during the video and the pre-test to identify each participant's average engagement; together with the calibration data, we calculated the smoothed baseline (via EMA) used for adaptive engagement classification in mode 3.

\subsection{Participants}

A total of 50 participants were recruited: 17 in the Full/unrestricted condition, 17 in Socratic condition, 16 in Adaptive condition. With 50 participants across three conditions, the study is exploratory in nature and detected moderate-to-large effects (Cohen's $d \geq 0.70$). Individuals were aged 18 to 40, recruited locally ($M = 27.82$, $SD = 5.34$). Of the 50 participants, 31 were female and 19 male (Full: 11F/6M; Socratic: 8F/9M; Adaptive: 12F/4M). 
The entire study was conducted in English, and all the materials used in the study were written in English. Participants were recruited from MIT campus and work or study in English : 26/50 were native English speakers (Full: 11/17, Socratic: 6/17, Adaptive: 9/16).

Participants were heavy AI users: 30/50 reported using chatbots daily and 12 weekly. 36/50 reported using them for learning, and the mean self-rated prompting skill was 3.1/5, with similar distributions across the three conditions. Self-reported prior familiarity with the topic of nuclear safety was very low ($M = 1.5/10$, median $= 1$, max $= 4$, no difference across conditions). Interest in the topic itself was not measured.

The study protocol was approved by the IRB of MIT (protocol ID 21070000428). Each participant was randomly assigned to exactly one mode via sequential random allocation: participant 1 to mode 1, participant 2 to mode 2, participant 3 to mode 3, participant 4 to mode 1, and so on.

\subsection{Materials}

We custom-built a web platform hosted on Netlify, using the OpenAI API with GPT-5.2.

\textbf{Instructional domain.} The chosen subject was nuclear safety protocols in a power plants, including defense-in-depth barriers, protection priorities, the chain of responsibility, and analysis of historical failures (Fukushima, Chernobyl). We chose this topic deliberately as a zero-prior-knowledge baseline; the background questionnaire included a self-assessment of prior familiarity to control for any unexpected baseline knowledge.

\textbf{Knowledge acquisition.} 10-minute instructional video with explanatory diagrams, developed from two IAEA publications \cite{iaea_design, iaea_safety}. Playback control and speed settings were disabled to ensure uniform exposure. The video covers all ten topics of the tests, but at a conceptual level: full credit on the test questions required a degree of details that the video alone rarely provided.

\textbf{Question design.} Three sets of 10 open-ended, short-answer questions (pre-test, training, post-test), for a total of 30 questions inspired by the IAEA publications \cite{iaea_design, iaea_safety}. Each set covered the same 10 conceptual topics, phrased differently across the 3 phases, enabling per-topic learning analysis.

\textbf{Questionnaires.} A background questionnaire captured demographics, sleep/alertness/physiological state, prior chatbot experience and use patterns, as well as assessment of prior knowledge on the topic of the study. A post-experiment questionnaire, tailored to each mode (3 slightly different versions), shared a common core: perceived learning, problem difficulty, enjoyment, self-reported engagement trajectory, and mode-specific items.

\textbf{EEG hardware.} The Muse 2 headband \cite{muse2} with 4 active channels (AF7, AF8 on the forehead; TP9, TP10 behind the ears), with Fpz as the reference electrode, mapped to the international 10--20 system. EEG data streams at 256 Hz via Bluetooth and was worn throughout the entire experiment for all 3 Modes.

\subsection{EEG Signal Processing Pipeline}

Raw EEG data is acquired in the browser using \texttt{muse-js} library and processed in real time with \texttt{rxjs} with \texttt{@neurosity/pipes}. The processing pipeline follows established methods from Neurochat paper \cite{baradari2025}.

The pipeline proceeded as follows: (1)~\textit{Sample alignment} of the raw samples from all 4 channels using \texttt{zipSamples}; (2)~\textit{Filtering} with a band-pass filter (1--30 Hz) and a 60 Hz notch filter to remove power line interference; (3)~\textit{Epoching} into 1-second epochs (250 ms overlapping during calibration, non-overlapping during the experiment); (4)~\textit{Spectral decomposition} via FFT into Theta (4--8 Hz), Alpha (8--13 Hz), and Beta (13--30 Hz); (5)~\textit{Engagement computation} $E = \text{Beta}/(\text{Alpha} + \text{Theta})$, introduced by Pope et al. \cite{pope1995}, where higher values indicate greater cognitive engagement; (6)~\textit{Normalization} of $E$ to $E_{\text{norm}} \in [0, 1]$ using the calibration $E_{\min}$ and $E_{\max}$: $E_{\text{norm}} = (E - E_{\min})/(E_{\max} - E_{\min})$.

For mode 3, the participant's engagement is classified in real time into three levels (high, normal, low), which directly drives the adaptive visual feedback and question reformulation. The initial baseline is the equally weighted average of the calibration, pre-test, and video phase's mean engagement, and is gradually updated after each problem using an Exponential Moving Average (EMA) with a small weight, so the baseline drifts slowly while still adapting to natural shifts in cognitive state.

\subsection{Grading and Evaluation}

\textbf{Real-time grading (assessment phase).} 
During the training phase (for Modes 1, 2, and 3), answers were graded in real-time by GPT-5.2, using a standardized prompt that includes the problem statement, the student's answer, the correction and the notation grid. The temperature was set to 0 for deterministic outputs, so with the fixed rubric the same answer almost always receives the same score. We observed no participant blocked by a hallucinated grade; the complaint reported by some participants was rather that reaching 100\% required specific vocabulary, whereas the 80\% pass threshold indicated conceptual understanding. While automated grading may not be perfectly accurate in absolute terms, it was applied in the same way across all Modes with the same parameters. We acknowledge, however, that this does not by itself guarantee identical grading behavior across conditions --- in particular, GPT may grade GPT-phrased answers more favorably; Section~\ref{sec:similarity} quantifies this concern and we discuss this further in the Limitations section.

\textbf{Pre-test and post-test grading.} All pre- and post- test responses were graded by GPT-5.2 with the exact same parameters to maintain consistent criteria. The scoring grid was: 90--100 (conceptually complete and correct), 70--89 (mostly correct but not the specific vocabulary), 40--69 (partial understanding, misses major elements); 1--39 (mostly incorrect but not blank); 0 (blank or irrelevant). Temperature was set to 0 so identical answers always receive the same score. During the sessions, a human evaluator informally spot-checked GPT scores (1--2 questions per participant, 60--120 items total) to make sure they were reasonable; however, those checks were never written down as independent scores, so no human--AI inter-rater reliability statistic can be computed retrospectively (see Section~\ref{sec:limitations}).

%% file: sections/data_analysis_plan.tex
\subsection{Primary and Secondary Metrics}

To measure the impact of each learning approach, we looked at several different metrics:
\begin{itemize}
    \item \textbf{Primary Metric : Learning Delta ($\Delta$):} The difference between each post-test and pre-test question's scores (across the 10 conceptual topics). This is the main dependent variable for between-group comparisons.
    \item \textbf{Primary Metric : Cognitive Engagement:} The normalized EEG Engagement ($E_{\text{norm}}$) measured across each experimental phase and per question, compared between modes in the training phase.
    \item \textbf{Secondary Metrics:}
    \begin{itemize}
       \item comparison of the interactions with the restricted pedagogical bot and the unrestricted Bots
       \item Number of interactions/attempts required to reach the 80\% mastery threshold
        \item Self-reported engagement and perceived learning (from post-experiment questionnaires).
    \end{itemize}
\end{itemize}

\subsection{Statistical Methods}

Two statistical approaches were applied to each primary metric: a \textbf{one-way ANOVA} on participants' average learning Delta and engagement, and a \textbf{Linear Mixed Model} (LMM) on the full per-question repeated-measures data. The ANOVA tested whether the three mode group means differed significantly, using each participant's mean score (averaged across the 10 questions) as a single observation ($N = 50$). The LMM addressed the hierarchical structure of the data (questions clustered within participants, within modes) by modeling all $N = 500$ per-question observations simultaneously, with modes as a fixed effect and participant as a random intercept:
\[
Y_{ij} \sim \text{Mode}_j + (1 \mid \text{Participant}_i)
\]
where $Y_{ij}$ is the dependent variable (learning delta $\Delta_q$ or normalized engagement $E_{\text{norm}}$) for question $q$ of participant $i$ in mode $j$.

\paragraph{ANOVA computation.} When the omnibus $F$-test is significant ($p < .05$), complementary comparisons are conducted using Welch's $t$-tests, with Cohen's $d$ as the effect size.

\paragraph{LMM computation.} The LMM was estimated using REML, with pairwise modes compared via estimated marginal means (we report $z$-statistic and 95\% CI). For learning gains the model ran on 500 change scores; for EEG engagement on all 22,775 cleaned EEG segments recorded during training.

%% file: sections/results.tex
All 50 participants completed the experiment. Results were compared based on four criteria described above : baseline equivalence, learning outcomes (learning delta), cognitive engagement (EEG), and conversational interaction patterns (discussion clusters).

\subsection{Baseline Equivalence: Pre-Test Scores}

A critical prerequisite for a valid group comparisons is that participants began the experiment with equivalent knowledge levels. The three groups exhibited highly similar mean pre-test scores: Full ($\bar{S}_{\text{pre}} = 37.19$, $SD = 8.33$), Socratic ($\bar{S}_{\text{pre}} = 38.40$, $SD = 8.30$), and Adaptive ($\bar{S}_{\text{pre}} = 34.49$, $SD = 8.49$). A one-way ANOVA confirmed no significant difference between groups ($F(2, 47) = 0.938$, $p = .399$). This shows that the three modes' groups started with approximately equivalent baseline knowledge of the nuclear safety domain, confirming the validity of the zero-prior-knowledge assumption and ensuring that any post-test differences reflect the effect of the training mode rather than pre-existing knowledge differences. 

\subsection{Learning Delta ($\Delta$)}

\subsubsection{Overall Learning Delta by Mode}

The overall mode delta is the mean across all 10 questions and all participants in that group, resulting in $N = 170$ observations for Full and Socratic (17 participants $\times$ 10 questions) and $N = 160$ for Adaptive (16 participants $\times$ 10 questions).

Table~\ref{tab:overall_delta} presents the aggregate learning delta statistics for each mode.

\begin{table}[htbp]
\centering
\caption{Learning delta ($\Delta$) statistics by Mode.}
\label{tab:overall_delta}
\begin{tabular}{lcccc}
\hline
\textbf{Mode}& \textbf{N} & \textbf{Mean $\Delta$} & \textbf{SD} & \textbf{Median} \\
\hline
Full (unrestricted) & 17 & 20.12 & 11.85 & 25.00 \\
Socratic      & 17 & 10.52 & 12.02 &  6.50 \\
Adaptive            & 16 & 11.20 &  8.95 &  9.55 \\
\hline
\end{tabular}
\end{table}

The Full (unrestricted) mode produced the highest overall learning delta ($\bar{\Delta} = 20.12$, $SD = 11.85$), compared to the Socratic mode ($\bar{\Delta} = 10.52$, $SD = 12.02$) and the Adaptive mode ($\bar{\Delta} = 11.20$, $SD = 8.95$) conditions. This result is against hypothesis H1, which predicted that the Socratic condition would result in higher learning gains than the unrestricted condition through productive struggle and deeper cognitive processing.

\paragraph{Statistical tests.} A one-way ANOVA on the per-participant mean delta revealed a significant main effect of mode ($F(2, 47) = 3.948$, $p = .026$). Pairwise Welch's $t$-tests with Cohen's $d$ effect sizes showed:

\begin{itemize}
    \item \textbf{Full vs.\ Socratic:} $t = 2.345$, $p = .025$, $d = 0.80$ (large effect). Participants with unrestricted access achieved significantly higher deltas than those in the Socratic condition.
    \item \textbf{Full vs.\ Adaptive:} $t = 2.448$, $p = .021$, $d = 0.85$ (large effect). The unrestricted condition also outperformed the adaptive exercise system.
    \item \textbf{Socratic vs.\ Adaptive:} $t = -0.186$, $p = .854$, $d = -0.06$ (negligible effect). The modalities 2 and 3 produced statistically indistinguishable learning outcomes.
\end{itemize}

\paragraph{Complementary LMM analysis.} To account for the repeated measures (10 learning delta per participant), we additionally fit a linear mixed model on every deltas ($n = 500$ rows):
\[
\Delta_{q} \sim \text{Mode} + (1 \mid \text{Participant}).
\]
The results were consistent with the ANOVA summary (Full: $20.12$, Socratic: $10.52$, Adaptive: $11.20$) :
\begin{itemize}
    \item \textbf{Socratic / Full:} estimate $=-9.60$, $SE=3.68$, $z=-2.61$, $p=.009$, 95\% CI $[-16.82, -2.38]$.
    \item \textbf{Adaptive / Full:} estimate $=-8.92$, $SE=3.74$, $z=-2.39$, $p=.017$, 95\% CI $[-16.25, -1.59]$.
    \item \textbf{Adaptive / Socratic:} estimate $=0.68$, $SE=3.74$, $z=0.18$, $p=.855$, 95\% CI $[-6.65, 8.01]$.
\end{itemize}
The LMM confirms the same pattern as the ANOVA/Welch analyses : the Full condition significantly outperforms the 2 other conditions, while Socratic and Adaptive do not differ significantly.

\subsubsection{Per-Question Learning Delta}

Examining the mean delta per question by mode, the Full condition achieved the highest delta on most questions (Q1 to Q7), with particularly strong gains on Q4 ($\Delta = 31.53$) and Q7 ($\Delta = 27.76$). The Adaptive mode showed a progressive increase in delta across the session (from $\Delta = 3.16$ on Q1 to $\Delta = 19.97$ on Q9), suggesting that participants benefited increasingly from the adaptive mode as they got to understand the system. The Socratic mode exhibited more variable performance.

\subsection{EEG Cognitive Engagement}

\subsubsection{Engagement During the Training Phase}

The three groups exhibited comparable engagement during all pre-training phases: calibration (Full: $0.546$, Socratic: $0.522$, Adaptive: $0.590$), video (Full: $0.519$, Socratic: $0.477$, Adaptive: $0.477$), and pre-test (Full: $0.436$, Socratic: $0.447$, Adaptive: $0.464$). A one-way ANOVA on baseline engagement (mean across calibration, video, and pre-test) confirmed no significant group differences for the pre training phases ($F(2, 47) = 0.392$, $p = .678$), validating the between-group comparison on the training phase. Table~\ref{tab:eeg_mode} reports the detailed statistics for the training phase.

\begin{table}[htbp]
\centering
\caption{Mean EEG engagement index ($E_{\text{norm}}$) during training by mode.}
\label{tab:eeg_mode}
\begin{tabular}{lcccc}
\hline
\textbf{Mode}& \textbf{N} & \textbf{Mean $E_{\text{norm}}$} & \textbf{SD} & \textbf{Median} \\
\hline
Full (unrestricted) & 17 & 0.389 & 0.129 & 0.395 \\
Socratic      & 17 & 0.460 & 0.211 & 0.422 \\
Adaptive            & 16 & 0.547 & 0.215 & 0.519 \\
\hline
\end{tabular}
\end{table}

Only the Adaptive condition's mean exceeds the 0.5 baseline threshold, indicating that participants in this condition were, on average, more cognitively engaged during the training phase than during the preceding calibration, video, and pre-test phases.

\paragraph{Statistical tests.} A one-way ANOVA on per-participant mean $E_{\text{norm}}$ during training resulted in a marginally significant main effect of mode ($F(2, 47) = 2.908$, $p = .064$). Pairwise Welch's $t$-tests revealed:

\begin{itemize}
    \item \textbf{Full vs.\ Adaptive:} $t = -2.545$, $p = .018$. The Adaptive condition produced significantly higher engagement than the unrestricted chatbot.
    \item \textbf{Socratic vs.\ Adaptive:} $t = -1.170$, $p = .251$. The difference between Socratic and Adaptive engagement did not reach significance.
    \item \textbf{Full vs.\ Socratic:} $t = -1.192$, $p = .244$. No significant difference between the two conversational modalities.
\end{itemize}

\paragraph{Complementary LMM analysis.} We also fit a mixed model at the EEG epoch level during training ($n = 22{,}775$ epochs): 
\[
E_{\text{norm}} \sim \text{Mode} + (1 \mid \text{Participant}).
\]
Estimated means were Full $=0.388$, Socratic $=0.460$, Adaptive $=0.547$. Pairwise contrasts showed:
\begin{itemize}
    \item \textbf{Socratic - Full:} estimate $=0.0716$, $SE=0.0627$, $z=1.14$, $p=.253$, 95\% CI $[-0.051, 0.195]$.
    \item \textbf{Adaptive - Full:} estimate $=0.1588$, $SE=0.0637$, $z=2.49$, $p=.013$, 95\% CI $[0.034, 0.284]$.
    \item \textbf{Adaptive - Socratic:} estimate $=0.0871$, $SE=0.0636$, $z=1.37$, $p=.171$, 95\% CI $[-0.038, 0.212]$.
\end{itemize}
This complementary measures analysis confirms the ANOVA interpretation: participants in the Adaptive mode have an engagement that is significantly higher than the Full mode, while differences involving Socratic remain non-significant.

These results support hypothesis H2: the Adaptive condition significantly outperformed the Full condition in engagement and focus during the task.

\subsubsection{Per-Question Engagement}

The Adaptive and Full mode maintained relatively stable engagement throughout the session. The Socratic mode exhibited its highest engagement in the first half of the session (Q2 to Q5, reaching $0.529$ on Q5) followed by a decline toward the final questions (Q10: $E_{\text{norm}} = 0.297$).
This late-session disengagement in the Socratic condition is further analyzed in Section~\ref{sec:discussion_clusters}.

\subsection{Discussion Strategy Analysis (Mode 1 vs.\ Mode 2)}
\label{sec:discussion_clusters}

\subsubsection{Computation Methodology}

Although Modes 1 and 2 share the exact same interface and model, the two conditions diverged on every behavioral measure we collected: median training time, number of attempts, copy-pasting, and perceived difficulty. Time on task differed as expected by design, with Socratic sessions taking longer than Full sessions.

To understand \textit{how} participants interacted with the conversational AI, we developed a rule-based classification system that categorizes each discussion into one of eight behavioral clusters. This analysis applied only to mode 1 (Full/unrestricted) and mode 2 (Socratic), as mode 3 (Adaptive) has no conversational interface. For each exercise attempt (defined as the sequence of user messages and AI responses before submission), we first detected copy-paste of the exercise statement or of the chatbot's answer. If the message was not a copy-paste, user messages were matched against pattern-based categories: \textit{ask for explanation} (e.g.\ ``why'', ``explain''), \textit{express confusion} (``I don't understand'', ``not sure''), \textit{ask for definition} (``what is/are'', ``define''), \textit{ask for confirmation} (``is this correct'', ``am I right''), and \textit{propose/refine answer} (``I think'', ``maybe...?'', ``let me try''). Discussions with no chatbot interaction were labeled \textit{No discussion} and those matching no category as \textit{Other}. Each classified discussion was linked to the training score for that exercise.

The categories were derived from our pilot sessions and match the classic help-seeking distinction between getting the answer (copy-paste) and trying to gain an understanding (explanation, definition, confusion, proposing an answer) \cite{nelsonlegall1981,aleven2003}. The rules were applied (copy-paste detection first, then the pattern-based categories), so each discussion received a single dominant label even when a conversation mixed several strategies. The \textit{Other} category is mostly composed of content questions (e.g., ``Which are the barriers in a nuclear system?''), not of off-topic behavior.

\subsubsection{Distribution of Discussion categories}

Table~\ref{tab:clusters} summarizes the proportional distribution of each discussion category across the two conversational modes:

\begin{table}[htbp]
\centering
\caption{Discussion strategy distribution by mode.}
\label{tab:clusters}
\resizebox{\columnwidth}{!}{%
\begin{tabular}{l|rr|rr}
\hline
\textbf{Strategy} & \textbf{Full} & \textbf{\%} & \textbf{Socratic} & \textbf{\%} \\
\hline
Copy-paste question    & 90 & 52.9 & 37 & 21.8 \\
Ask for definition     & 38 & 22.4 & 49 & 28.8 \\
No discussion          & 26 & 15.3 & 38 & 22.4 \\
Other                  & 11 &  6.5 & 28 & 16.5 \\
Ask for explanation    &  3 &  1.8 &  3 &  1.8 \\
Express confusion      &  2 &  1.2 & 11 &  6.5 \\
Propose/refine answer  &  0 &  0.0 &  4 &  2.4 \\
Ask for confirmation   &  0 &  0.0 &  0 &  0.0 \\
\hline
\textbf{Total}         & \textbf{170} & & \textbf{170} & \\
\hline
\end{tabular}%
}
\end{table}

\subsubsection{Behavioral Observations During the Study}

Beyond the automated cluster analysis, direct observation of participants during the experiment sessions revealed distinct behavioral patterns in each conversational mode.

\paragraph{Mode 1 (Full/unrestricted).} The majority of Mode 1 participants quickly discovered that the chatbot could provide direct and complete answers to the exercises. Once this was understood, most participants shifted their strategy toward systematic answer extraction: they would copy-paste the exercise prompt, retrieve the chatbot's response, and transcribe it as their answer. Some participants also used the chatbot to reformulate or simplify the question before requesting the answer. The data clustering confirms this behavior : 52.9\% of all exercise interactions in Mode 1 were classified as \textit{copy-paste question}, and the remaining interactions were mostly \textit{ask for definition} (22.4\%).

\paragraph{Mode 2 (Socratic).} In Mode 2, where the chatbot is constrained to never provide the final answer and instead guides through questions, two distinct behavioral profiles emerged among participants. The first and larger group quickly got frustrated with the constraints of the Socratic approach. These participants approached the chatbot with the same mental model as a general-purpose LLM : expecting direct, immediate answers. When the chatbot consistently redirected them by asking guiding questions or providing guidance instead of direct solutions, these participants progressively disengaged from the chatbot, using it less and less as the session progressed.
The second, smaller group of participants genuinely engaged with the Socratic dialogue: they asked follow-up questions and explored concepts in depth.

\subsubsection{Quantitative Evidence: Chatbot Disengagement Over Time}

Mode 2 participants exchanged far more messages overall ($\bar{M} = 119.2$ total messages per session, $SD = 50.2$) compared to Mode 1 ($\bar{M} = 37.3$, $SD = 19.8$). This is expected: the Socratic chatbot's refusal to provide direct answers forces participants into longer exchanges to arrive at the solution. However, the per-question trajectory reveals a disengagement pattern in Mode 2 (the mean number of chat messages declines after Q5). In Mode 1, the decline is also present but more moderate, reflecting a natural efficiency gain as participants learned to talk with the chatbot.

This behavioral disengagement in Mode 2 converges with the EEG data: the Socratic mode's EEG engagement drops to its lowest point at Q10 ($E_{\text{norm}} = 0.297$, well below the 0.5 baseline). Together, these two independent measures , behavioral (fewer messages) and neurophysiological (lower engagement), indicate that the majority of Mode 2 participants who expected the chatbot to talk as an answer-providing tool progressively abandoned it when it failed to meet their expectations.

\subsubsection{Linguistic Similarity Between Participant Answers and Chatbot Text}
\label{sec:similarity}

To quantify how much participants' submitted answers were contaminated by the chatbot's phrasing, we compared each final training answer with the chatbot messages of the corresponding exercise. Mean similarity between participant answers and chatbot messages was 0.374 in Mode 1 versus 0.044 in Mode 2 ($p < 10^{-18}$), and 25.3\% of Mode-1 final training answers were nearly identical to the chatbot text, against 0\% in Mode 2. This confirms that Mode 1 answers were heavily derived from the chatbot, while Mode 2 answers were formulated by the participants themselves.

{
\subsection{Number of Attempts to Reach Mastery}

A key feature of the training phase is the 80\% threshold: participants must achieve a score of at least 80 on each exercise before moving on to the next question. Socratic participants required nearly twice as many attempts ($M = 2.80$, $SD = 0.73$) as Full participants ($M = 1.46$, $SD = 0.59$); Welch's $t = -5.865$, $p < .0001$, $d = -2.01$ (very large effect). The comparison excludes Mode~3, where the adaptive condition redirects participants to simpler sub-questions after an incorrect answer. The threshold was almost never reached in the Socratic condition: 58\% of the exercises of participants in Socratic Mode, who abandoned the dialogue mid-way, never reached the 80\% threshold, meaning those participants ended the exercise without having produced a complete, correct formulation of the answer. We saw no evidence of participants being blocked by a hallucinated grade, but sometime the barrier was the specific vocabulary required for a full credit.

\subsection{Self-Reported Engagement and Perceived Learning}

To complement the objective measures of learning and engagement, the post-questionnaire asked participants to rate their subjective experience on several points. These self-report measures address hypothesis H3.

\subsubsection{Perceived Learning}

Participants rated how much they thought they had learned during the session on a 0-10 scale. The three modes produced very similar scores (Full: $M = 6.06$, $SD = 1.81$; Socratic: $M = 6.11$, $SD = 2.03$; Adaptive: $M = 5.75$, $SD = 1.95$). A one-way ANOVA revealed no significant difference between groups ($F(2, 47) = 0.169$, $p = .845$). Participants in Full mode, who achieved nearly double the learning delta of the other two conditions, did not perceive themselves as having learned more.

\subsubsection{Perceived Difficulty}

Participants rated the difficulty of the questions on a 1-10 scale (Full: $M = 6.75$, $SD = 1.95$; Socratic: $M = 8.06$, $SD = 1.55$; Adaptive: $M = 7.88$, $SD = 1.71$). The omnibus ANOVA was marginally significant ($F(2, 47) = 2.734$, $p = .075$). Pairwise comparisons indicated that Socratic participants perceived the exercises as significantly more difficult than Full participants ($t = -2.143$, $p = .041$, $d = -0.74$), while the Adaptive-Full comparison was marginally significant ($t = -1.736$, $p = .093$, $d = -0.61$). No significant  difference between Socratic and Adaptive ($p = .750$).
These results are consistent with the objective data: the two constrained modes (mode 2 and 3) imposed greater cognitive demands, while the copy-paste strategy in mode 1 made the exercises feel easier.

Perceived difficulty was not correlated with objective engagement: per participant, perceived difficulty and mean EEG engagement during training showed no correlation ($r = .126$, $p = .388$, $n = 49$), with the same null pattern within each condition. Perceived difficulty did not correlate with perceived learning either ($r = -.06$).

\subsubsection{Chatbot and System Helpfulness}

For the two conversational modes, participants rated the chatbot's helpfulness on a 0-10 scale. The Full condition rated the chatbot as highly helpful ($\bar{x} = 8.62$, $SD = 1.31$), while the Socratic condition rated it notably lower ($\bar{x} = 5.67$, $SD = 2.00$); Welch's $t = 5.154$, $p < .0001$, $d = 1.75$ (very large effect). Several Socratic participants explicitly described their frustration: ``\textit{It felt annoying and frustrating}'' (P035); ``\textit{Feeling almost crazy}'' (P029); ``\textit{I just gave up}'' (P014).

}

%% file: sections/discussion.tex
\subsection{Interpretation of Results}

The central finding is that the unrestricted chatbot (mode 1) produced nearly double the learning delta of the Socratic (mode 2) and Adaptive (mode 3) conditions, going against hypothesis H1. However, the data and analysis of the bot discussions suggests that the success of the unrestricted AI is not evidence of deeper learning, but rather a result of the immediate post-test evaluation after the training phase.

\paragraph{Direct answer retrieval in Mode 1.} The discussion cluster analysis (Table~\ref{tab:clusters}) provides an explanation for mode 1's better learning delta: 52.9\% of all exercise interactions were classified as \textit{copy-paste question}, meaning that participants pasted the exercise statement directly into the chatbot and received the complete answer. Because the post-test was administered immediately after the homework phase, this short-term retention was sufficient to produce high scores. The per-question delta supports this interpretation: mode 1 consistently achieved the highest deltas on Q1 to Q7, precisely the questions where participants spent the most energy extracting answers from the chatbot.

\paragraph{Participants ``gave up'' on the Socratic chatbot.} In contrast, mode 2 participants faced a very different interaction dynamic. The Socratic chatbot's design to refuse providing final answers forced participants into extended dialogues. While this is pedagogically sound in principle, especially with respect to supporting a productive struggle and deeper understanding, the data reveals that most of the participants did not complete this process and "gave up". \textit{Behaviorally}, the mean number of messages per question in mode 2 declined after question~5, showing that participants progressively stopped engaging with the chatbot's guided questioning. \textit{Neurophysiologically}, EEG engagement in mode 2 dropped to its lowest value at Q10 ($E_{\text{norm}} = 0.297$), well below the 0.5 personal baseline (Full: $E_{\text{norm}} = 0.451$; Adaptive: $E_{\text{norm}} = 0.443$).

Participants who abandoned the Socratic dialogue before reaching a complete understanding were left with only a partial or fragmented answer to retain for the post-test, penalizing performance regardless of any partial understanding developed.

\paragraph{Why the shared video did not provide equal outcomes.} The video was complete but not sufficient: right after watching it, all groups scored only 34--38/100 on the pre-test (no group difference, $p = .399$). Post-test performance therefore mostly reflects the training phase. Mode 1 repeatedly showed complete, well-formulated answers (52.9\% copy-paste). Socratic participants who quit mid-dialogue never got a complete formulation at all (58\% of their responses never reached the threshold): little to retain, whatever partial understanding they had built.

Seen through the lens of cognitive load theory \cite{sweller1998,sweller2010}, the three conditions distributed mental effort very differently: Mode 1 let participants finish the task with almost no mental effort, Mode 2 required an effort that most participants eventually gave up on, and Mode 3 kept engagement by splitting the task into sub-questions itself, taking over some of the load.

\paragraph{Immediate testing favors retention of answers over understanding.} The post-test followed the assessment session with no delay, so the experiment measures short-term recall rather than deep conceptual understanding or long-term retention. Mode 1's complete answers create an optimal pathway for immediate recall, mirroring the documented "illusion of competence,'' where exposure to complete solutions results in high immediate performance \cite{bjork2011}. Mode 2's Socratic approach, by contrast, builds understanding through guided discovery, a process that takes time and effort.

\paragraph{Engagement and learning dissociation.} An important secondary finding is the dissociation between cognitive engagement (EEG) and learning outcomes. The Adaptive condition produced significantly higher EEG engagement than Mode 1 ($p = .018$), partially supporting H2, yet this did not translate into superior learning deltas ($\bar{\Delta} = 11.20$ for Adaptive vs.\ $20.12$ for Full). Cognitive engagement, while necessary, is not sufficient when the assessment rewards the completeness of the final answer rather than the depth of the process: mode 1 participants achieved high deltas with low engagement by transcribing AI answers, while mode 3 participants stayed focused but produced only moderate gains.

\paragraph{A null result for perceived learning.} Perhaps the most unexpected finding is the absence of any significant difference in perceived learning across modes ($p = .845$). Full mode participants, who objectively have a learning delta nearly twice bigger, did not feel they had learned more than Socratic or Adaptive mode participants. Full mode participants may have been aware that their good results were driven by answer retrieval rather than deep understanding (Quote from P031: ``\textit{A bit like cheating, copy/pasting context, questions and answers}''). The frustration-driven disengagement observed in the Socratic condition is a critical challenge for Socratic AI systems: this pedagogical path can only be beneficial if learners persist through the discomfort, which many did not.

\subsection{Practical Implications}

The choice of AI architecture must be aligned with the assessment format and learning timeline. Unrestricted AI access maximizes immediate post-session performance but may favor superficial knowledge; Adaptive approaches may be preferable for deeper understanding but must be carefully designed to sustain engagement and prevent premature abandonment. The Adaptive condition's significantly higher engagement demonstrates the promise of real-time neurophysiological feedback, but the translation of this engagement into measurable learning gains requires further work, particularly with delayed post-tests.

More specifically, our findings suggest one such design, in which each element is based on a finding of this study: (1) deployment of Socratic dialogue by default, but when EEG shows sustained disengagement below the personal baseline, switch temporarily to Mode-3-style sub-questions : the format that kept engagement highest, and then return to dialogue. (2) When struggle stops being productive (repeated failures, shrinking messages), give progressively more concrete hints instead of indefinite refusal, keeping the learner in the zone of proximal development (ZPD) \cite{vygotsky1978}. Such a system design is a suggestion, and should be evaluated in a formal study with a delayed post-test.

\subsection{Limitations and Future Work}
\label{sec:limitations}

Grading validity comes first among our study's limitations: the human checks of GPT scores during the sessions were done by eye to make sure the scores were reasonable, but they were never written down as scores, so we cannot compare human and GPT grading directly, and no inter-rater reliability can be reported; a systematic double-grading of a subset is planned as future work. It is worth noting, however, that the main behavioral results (copy-paste rates, disengagement, time on task, EEG) do not come from GPT scores, and they compliment the obrained results.

Our study was restricted to a single subject domain and mostly focused on factual knowledge acquisition, so replication across multiple domains and open-ended problem-solving questions is needed. Due to the Hawthorne effect \cite{mccambridge2014}, participants may behave differently because of the EEG device. The study accounts for short-term retention only; long-term retention was not measured, and the recommended 30-minute duration may have created implicit time pressure penalizing Mode 2 participants. With 16--17 participants per condition, the sample does not provide the power for subgroup analyses by AI/LLM proficiency use or age, which remain exploratory.

Additional limitations should be noted. Interest in the topic was not measured: nuclear safety was chosen as a zero-prior-knowledge domain, and we cannot rule out that low interest made the task feel like something to complete rather than an occasion to learn. Finally, Mode 3 participants had no hands-on familiarization phase with the EEG-triggered feedback, which likely explains the warm-up effect we observe (Mode 3's per-question delta climbs from 3.2 on Q1 to 20.0 on Q9 as participants learn the system): future versions of the studies of this nature should include a short practice session.

Two directions for future work emerge: a \textit{delayed post-test} (one to two weeks after the learning session) to analyze whether Mode 2 and Mode 3's deeper cognitive engagement translates into superior long-term retention; and replication with larger, more diverse participant pools, additional LLM models, and a human teacher control.

%% file: sections/conclusion.tex
This study compared three AI modes with 50 participants learning nuclear safety protocols. The unrestricted chatbot produced the highest learning delta ($\bar{\Delta} = 20.12$), outperforming both Socratic ($d = 0.80$) and Adaptive ($d = 0.85$) conditions. The Adaptive condition produced the highest EEG engagement ($p = .018$) and was the only condition above the 0.5 personal baseline. More than 50\% of mode 1 interactions involved direct answer retrieval, while mode 2 exhibited progressive disengagement. Perceived learning and perceived difficulty are self-reported measures and should be interpreted as such.

%% file: sections/appendix.tex
% ===== NEW APPENDIX (added in this revision) =====

\section{Socratic Tutor System Prompt}
\label{app:prompt}

\begin{footnotesize}
\begin{verbatim}
You are an encouraging nuclear technical protocols
training assistant.
You help students solve exercises without ever giving
the solution, even partially.
You have access to:
  - exerciseStatement = the exercise the student must
    solve
  - studentAnswer = the student's current attempt
  - correction = the reference solution (for your eyes
    only; NEVER reveal it)

Your goals:
  1. Guide the student step-by-step using questions,
     hints, and reasoning prompts.
  2. NEVER reveal any element of the correction, not
     even small steps.
  3. Do not solve the exercise for the student.
  4. Ask one question at a time, always based on what
     the student answered.
  5. Identify misconceptions by comparing studentAnswer
     with correction, but only use that to:
     - ask targeted questions
     - propose analogies
     - suggest a path of reflection
  6. If the student is wrong or stuck, give a hint,
     not a solution.
  7. Encourage the student to explain their thinking
     or try a smaller sub-step.
  8. When the student makes progress, praise and
     reinforce.
  9. When the student reaches the correct reasoning
     (even if imperfect), help them consolidate by
     asking them to restate the idea in their own
     words.
  10. At NO moment should you provide:
     - the correct answer
     - the missing step
     - the final numerical result
     - a direct formula application that reveals the
       solution.
  11. You can give definitions of concepts if the
      student asks for them.
Always answer in an upbeat, motivating tone, like a
real tutor.

At the end of every response, add one short
directional hint (one sentence max) that helps the
student know what to check next, without revealing
the solution. Prefix it with "Hint:".

Here is the inputs :
- The exercise statement: {exerciseStatement}
- The correction: {correction}
- The trainee's current response: {studentAnswer}
\end{verbatim}
\end{footnotesize}